\documentclass[letterpaper, 10 pt, conference]{ieeeconf}  
\pdfoutput=1

\IEEEoverridecommandlockouts                              

\usepackage{basicmath}
\usepackage{graphicx}   
\usepackage{amsmath} 
\usepackage{amssymb}  
\usepackage{units}
\usepackage[normalem]{ulem}

\usepackage{color} 
\let\labelindent\relax
\usepackage{enumitem}

\newcommand{\Will}[1]{\textcolor{blue}{#1}}

\renewcommand{\Will}[1]{#1}
\renewcommand{\sout}[1]{\ignorespaces}
\title{Impact Invariant Control with Applications to Bipedal Locomotion}

\author{William Yang and Michael Posa 
	\thanks{This material is based upon work supported by the National Science Foundation Graduate Research Fellowship Program under Grant No. DGE-1845298}
	\thanks{The authors are with the GRASP Laboratory, University of Pennsylvania, Philadelphia, PA 19104, USA \{yangwill, posa\}@seas.upenn.edu}%
}

\begin{document}

\maketitle

\thispagestyle{empty}
\pagestyle{empty}

\begin{abstract}
	When legged robots impact their environment, they undergo large changes in their velocities in a small amount of time.
	Measuring and applying feedback to these velocities is challenging, and is further complicated due to uncertainty in the impact model and impact timing.
    This work proposes a general framework for adapting feedback control during impact by projecting the control objectives to a subspace that is invariant to the impact event.
    The resultant controller is robust to uncertainties in the impact event while maintaining maximum control authority over the impact invariant subspace.
    We demonstrate the utility of the projection on a walking controller for a planar five-link-biped and on a jumping controller for a compliant 3D bipedal robot, Cassie.
    The effectiveness of our method is shown to translate well on hardware.

\end{abstract}

\section{Introduction}
\label{sec:introduction}

Handling the making and breaking of contact lies at the core of controllers for legged robots.
Its role becomes increasingly important as the the field demands that our legged robots be capable of more agile motions.
However, current controllers for legged robots are incredibly sensitive to these impact events.
When a robot's foot makes contact with the world, the foot is brought instantaneously to a stop by a large contact impulse.
The presence of large contact forces and rapidly changing velocities hinders accurate state estimation.
Coupled with the poor predictive performance of our contact models \cite{halm2019modeling} \cite{fazeli2020fundamental} \cite{remy2017ambiguous}, this combination of large state uncertainty and poor models makes control especially difficult.

Roboticists have attempted to improve the robustness of legged robots to these impact events by addressing the reference trajectories as well as the controllers that track those trajectories.
For example, the open-loop swing-leg retraction policy has been shown to have inherent stability to varying terrain heights \cite{seyfarth2003swing}. 
Qualitatively similar motions were also found independently through robust trajectory optimization \cite{dai2012optimizing} \cite{green2020planning}.
While designing more robust trajectories shows promise, the challenge of designing controllers to track these often discontinuous trajectories still remains.

\begin{figure}[t!]
	\centering
	\includegraphics[width=0.48\textwidth]{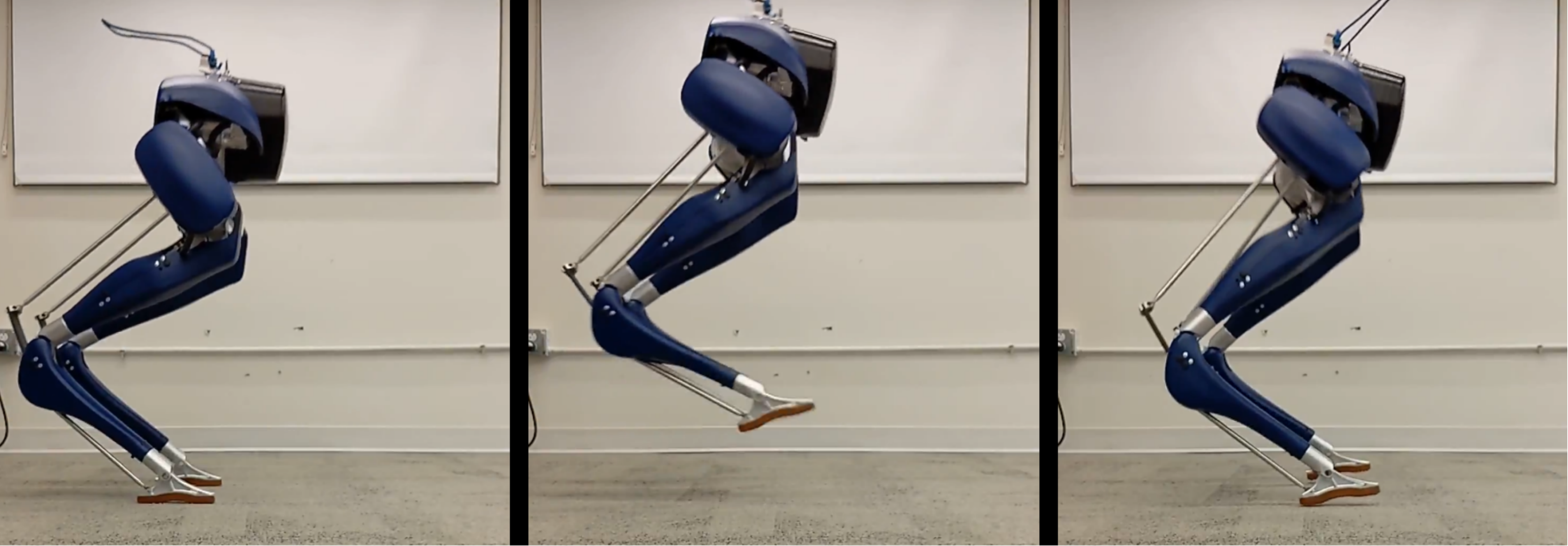}
	\caption{When legged robots such as Cassie execute agile high-impact motions like jumping, shown above, the joint velocities undergo rapid changes at the impact event. Attempting to apply feedback control on these velocities, using a reference trajectory that may be discontinuous due the rigid contact model assumption, is a difficult task. Often, the controller will actually introduce disturbances. This paper proposes a method to project both the robot velocities and the reference trajectory to a subspace that is invariant under impacts.}
	\label{fig:cassie_hardware_jump}
\end{figure}

Tracking a discontinuous trajectory is problematic due to the unavoidable difference between the reference trajectory and actual system caused by even \sout{minute} \Will{minuscule} differences in impact timing.
These differences cause feedback control efforts to spike, leading to instabilities. 
While these controller spikes can be reduced through strategies such as blending controller gains and contact constraints around the impact event \cite{mason2016balancing}, \cite{atkeson2015no}, these heuristic methods do not address the fundamental challenge of tracking discontinuous trajectories.
A strategy that does attempt to directly addresses this challenge is termed reference spreading control \cite{rijnen2017control}.
This method leverages contact detection and extending the reference trajectories to ensure that \Will{a valid reference trajectory exists despite mismatches in impact timing.}
\sout{ the tracking error does not jump when the reference trajectory anticipates an impact.}
\sout{However, the tracking error still jumps once when the actual impact occurs.}
\Will{However, during the transition between contact modes when the impact is still resolving, tracking even the extended reference trajectories can be detrimental.} 

Alternate methods, we note, focus instead on avoiding impact events altogether.
While impacts do not exist for frequently used templates such as the linear inverted pendulum (LIP) and the spring-loaded inverted pendulum (SLIP), impacts will manifest when embedding these templates onto physical robots with non-negligible mass in the legs.
Furthermore, it is neither possible nor desirable to avoid impacts for more agile motions such as running or jumping.
Thus, handling non-trivial impacts in a robust manner is essential to the development of more agile legged robots.

In this work, we propose a \sout{more principled} method for tracking discontinuous trajectories across impacts \Will{that directly avoids jumps in tracking error}. 
\sout{We avoid tracking jumps in tracking error, and thus avoid controller spikes, \textit{entirely}} 
\Will{We achieve this} by projecting the tracking objectives down to a subspace where they are invariant to the impact event.
\sout{ and consequently continuous. }
Gong and Grizzle \cite{gong2020angular} made an important insight about angular momentum about the contact point, noting that it is invariant to impacts at that contact point.
Inspired by this, we generalize this property and extend it to include the entire invariant subspace, which we term the impact invariant subspace.
The primary contribution of this paper is the identification of this subspace for the purposes of improving controller robustness to uncertainty in the impact event.
We develop a method for adapting controller feedback to be applied only on this subspace of velocities that are invariant to any contact impulses.
The subspace is easily defined for any legged robot at any given configuration, and the projection to that subspace can be applied to any tracking objective that is purely a function of the robot's state.
A key benefit of the impact invariant projection is \Will{that it enables controllers to be robust} \sout{robustness} to uncertainty in the impact event, \sout{while maintaining} \Will{while minimally sacrificing} control authority \sout{over the invariant subspace}.

To demonstrate the directional robustness to uncertainty in the impact event, we apply the projection to a walking gait for a planar five link biped.
Additionally we showcase the performance of the projection on an Operational Space Controller (OSC) tracking a jumping motion in simulation and on hardware for the bipedal robot Cassie.

\section{Background}
\label{sec:background}

\subsection{Rigid Body Dynamics}

	We use both the planar biped Rabbit \cite{chevallereau2003rabbit} and the 3D complaint bipedal robot Cassie to demonstrate the benefits of the idea of impact invariance.
	Both legged robots are modeled using conventional floating-base Lagrangian rigid body dynamics.
	Cassie has passive springs on its heel and knee joints; for the purposes of modeling and control we treat these springs as rigid.
	However, when evaluating our results in simulation, we do include these terms.

	The robot's state $x \in \Real^{2n} = \left[q; \dot q\right]$, described by its positions $q \in \Real^{n}$ and velocities $\dot q \in \Real^{n}$, is expressed in generalized floating-base coordinates\footnote{For notational simplicity, we assume $\dot q = v$, where $v$ are the velocities. For 3D orientation, as is the case for Cassie, this requires a straight-forward extension to use quaternions.}.
	The dynamics are derived using the Euler-Langrange equation and expressed in the form of the general manipulator equation:
	\begin{align}
		M(q) \ddot q + C(q, \dot q) + g(q) &= Bu + J_{\lambda}(q)^T \lambda,
		\label{eq:dynamics}
	\end{align}
	where $M$ is the mass matrix, $C$ and $g$ are the Coriolis and gravitational forces respectively, $B$ is the actuator matrix, $u$ is the vector of actuator inputs, and $J_{\lambda}$ and $\lambda$ are the Jacobian of the holonomic constraints and corresponding constraint forces respectively.

\subsection{Rigid Body Impacts}

	In this paper, we model the complex deformations and surface forces when a legged robot makes contact with a surface using a rigid body contact model.
	This contact model does not allow deformations; instead, impacts are resolved instantaneously.
	Therefore, the configuration remains constant over the impact event and the velocities change instantaneously according to the contact impulse $\Lambda$:
	\begin{align}
		M(\dot q^+ - \dot q^-) &= J_{\lambda}^T \Lambda,
		\label{eq:impact_map}
	\end{align}
	where $\dot q^+$ and $\dot q^-$ are the pre- and post-impact velocities, and $\Lambda$ is the impulse sustained over the impact event.
	With the addition of a standard constraint that the new stance foot does not move once in contact with the ground (no-slip condition):
	\begin{align}
		J_{\lambda} \dot q^+ = 0,
	\end{align}
	\noindent $\Lambda$ can be solved for explicitly, determining the post-impact state $x^+$ purely as a function of the pre-impact state $x^-$:
	\begin{align}
		q^+ &= q^- \label{eq:pos_reset_map},\\
		\dot q^+ &= (I - M^{-1} J_{\lambda}^T (J_{\lambda} M^{-1} J_{\lambda}^T)^{-1} J_{\lambda}) \dot q^- \label{eq:vel_reset_map}.
	\end{align}
	This reset map is conventionally enforced as a constraint between hybrid modes separated by an impact event for trajectory optimization of legged robots.
\section{Impact Invariance}
\label{sec:impact_invariance}

To motivate the concept of the impact invariant subspace, we begin by highlighting and describing the difficulties of applying feedback control during an impact event.
For the sake of simplicity, we consider a feedback controller with constant feedback gains that controls an output $y: \Real^n \rightarrow \Real^d$ to track a time-varying trajectory $y_{des}(t):~ \left[0, \infty \right) \rightarrow \Real^d$ by driving the tracking error $\tilde{y}(t) = ~y_{des}(t) - y(t)$ to zero.
This is commonly accomplished with a standard control law $u = u_{ff} + u_{fb}$ where $u_{ff}$ is the feedforward controller effort required to follow the reference acceleration $\ddot y_{des}$ and the $u_{fb}$ is the PD feedback component given by:
\begin{align}
	u_{fb}(t) = K_p \tilde{y}(t) + K_d \dot{\tilde{y}}(t).
\end{align}

\begin{figure}
	\centering
	\includegraphics[width=0.47\textwidth]{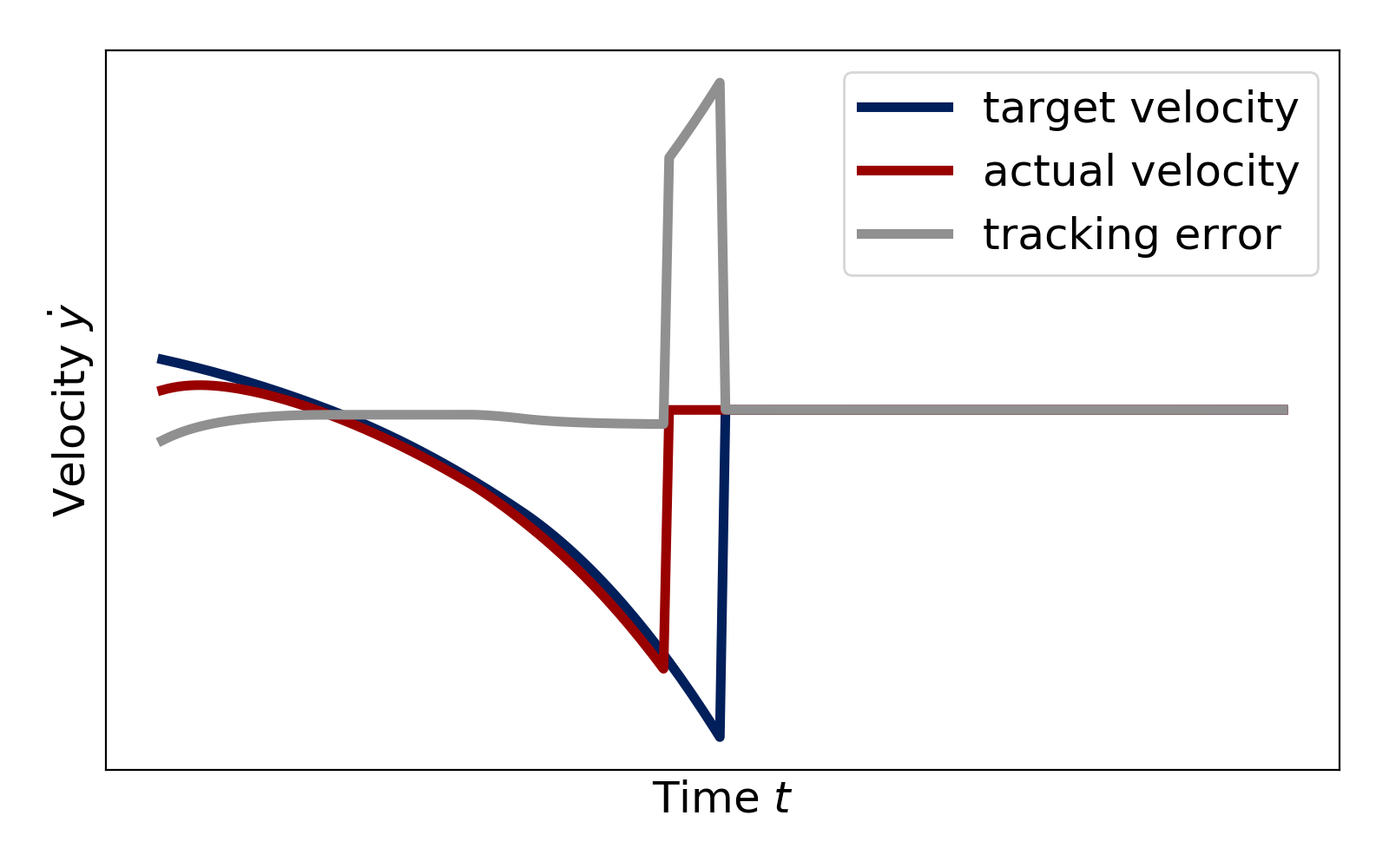}
	\caption{Illustration of a reference trajectory with a discontinuity due to an impact and a system that seeks to track that discontinuous trajectory. The velocity error will spike unless the actual and nominal impact time match up perfectly.}
	\label{fig:instantaneous_impacts}
\end{figure}

The reference trajectory $\dot{y}_{des}(t)$ for systems that make contact with their environment has discontinuities at the impact events in order to be dynamically consistent with \eqref{eq:vel_reset_map}.
Therefore, in a short time window around an impact event, there will be a discontinuity in the reference trajectory $\dot{y}_{des}(t)$ at the nominal impact time and another discontinuity when the actual system $\dot{y}(t)$ makes contact with the ground as shown in Fig. \ref{fig:instantaneous_impacts}.
Because the robot configuration is approximately constant over the impact event, the change in controller effort is governed by the change in velocity error:
\begin{align}
	\Delta u \approx K_d \Delta \dot y,
\end{align}
thus any mismatches in impact timings will unavoidably result in spikes in the feedback error and therefore control effort as similarly noted in \cite{rijnen2017control}.

\remark{\textit{We make an assumption that the jump in the reference trajectory is time-based. Although it is possible to formulate trajectories with event-triggered jumps, these methods require detection, which for state-of-the-art methods still have delays of 4-5ms \cite{bledt2018cheetah}. Moreover, in reality, impacts are not resolved instantaneously but rather over several milliseconds. In this time span, it is not clear which reference trajectory to use as using either trajectory will output a large tracking error.}}

Note that a large tracking error, shown in Fig. \ref{fig:instantaneous_impacts}, results from only a small difference in impact timing, yet the controller will respond to the large velocity error and introduce controller-induced disturbances. 
This sensitivity to the impact event is amplified by the large contact forces that impair state estimation and inaccuracies in our contact models \cite{fazeli2020fundamental}, meaning that the velocities post-impact may not match up with the reference trajectory.
Detecting the ``true" error for discontinuous functions is a difficult problem and has been explored in \cite{pfrommer2020contactnets}.

The key insight in resolving this problem is inspired by \cite{gong2020angular}, in which Gong and Grizzle delineate desirable properties of angular momentum about the contact point, denoted as $L$.
They highlight that $L$ is invariant over impacts on flat ground, meaning that it is continuous over the impact event despite it being a function of velocity.

The concept of an impact invariant subspace is a generalization of this property.
We observe that there is a space of velocities that, like $L$, are continuous through impacts for \textit{any} contact impulse.
By switching to track these outputs in a small time window around anticipated impacts, we avoid controller-induced disturbances from uncertainty in the impact event.
Note that while $L$ $\in \Real^3$ or $\Real^2$ for planar systems such as Rabbit, the impact invariant subspace $\in \Real^{n - c}$, where $n$ is the dimension of generalized velocities and $c$ is the number of independent constraints of the impact event.
For Rabbit, this space is $\in \Real^{7-2}$.
For Cassie, each foot provides 5 holonomic constraints and the four bar linkage on each leg provides 2 additional holonomic constraints that are always active.
Therefore the impact invariant subspace is $\in \Real^{18 - 7}$ for impacts with a single foot (walking, running) and $\in \Real^{18 - 12}$ for impacts with both feet (jumping).
A direct benefit of this higher dimensional space is the higher degree of possible control, which enables more agile or energetically efficient motions.

The impact invariant subspace is defined as the nullspace of $M^{-1} J_{\lambda}^T$, which is the matrix that maps contact impulses to generalized velocities.
Thus a basis $P(q) \in \Real^{(n - c) \times n}$ for this nullspace is such that:
	\begin{align}
		P (\dot q - \dot q^-) = 0 = P M^{-1} J_{\lambda}^T \Lambda.
		\label{eq:ii_projection}
	\end{align}
This creates the intended effect, that is, for any contact impulse $\Lambda$, the impact invariant velocities will be unchanged.
Alternatively, to project the generalized velocities down to the impact invariant subspace, we can simply create a low-rank $\Real^{n \times n}$ projection matrix $Q(q) = P^T P$.

\begin{figure}
	\centering
	\includegraphics[width=0.48\textwidth]{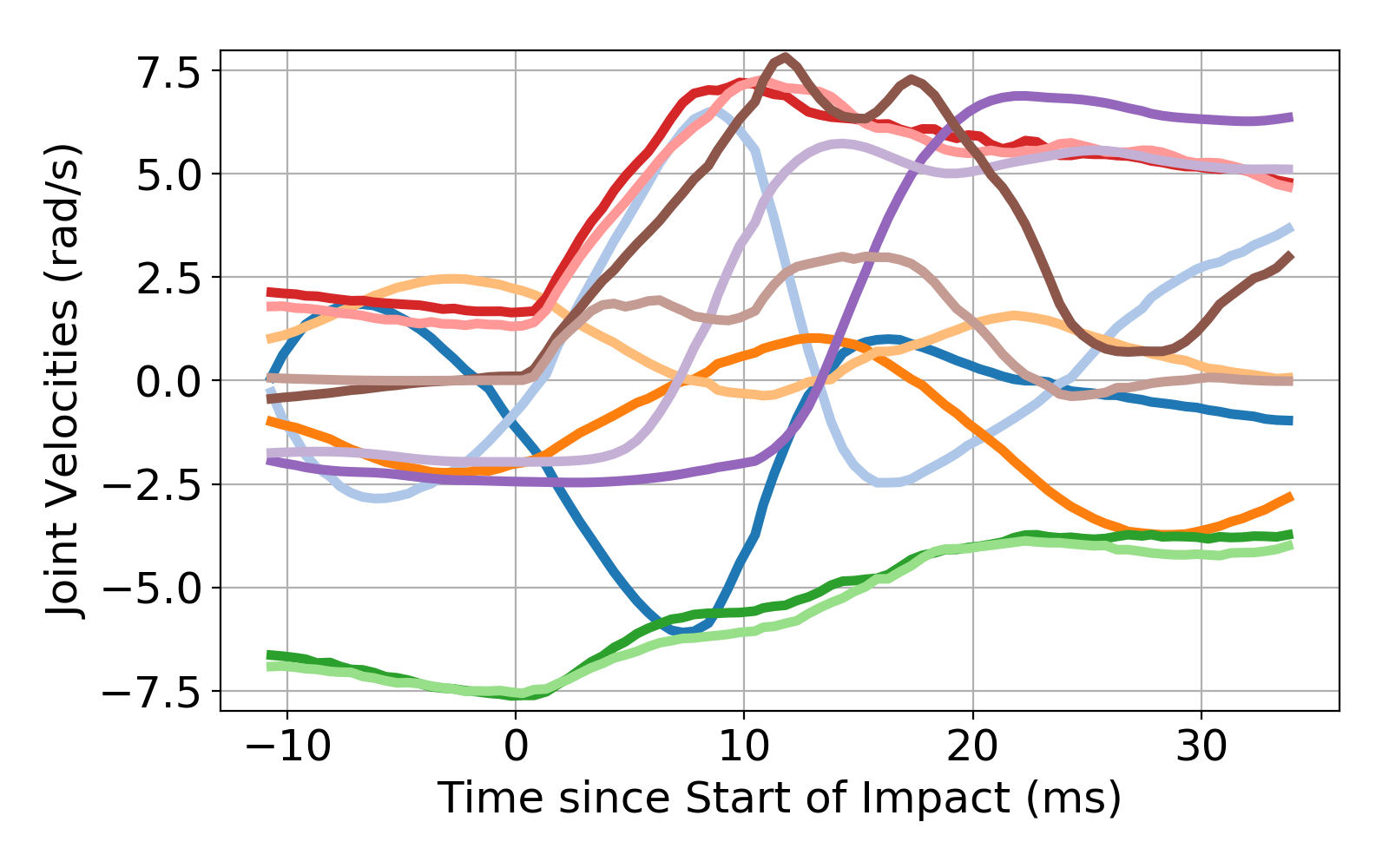}
	\includegraphics[width=0.48\textwidth]{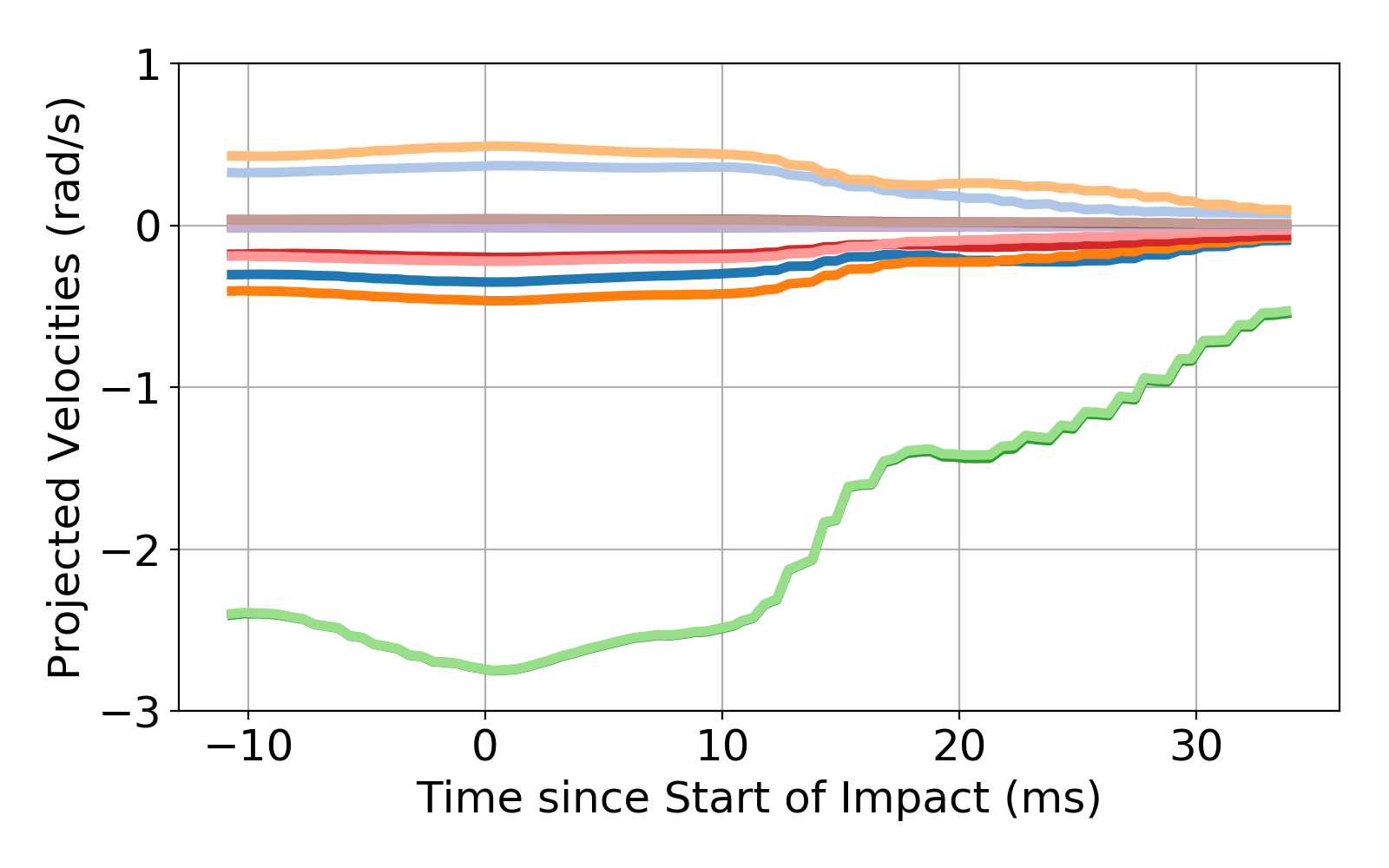}
	\includegraphics[width=0.32\textwidth]{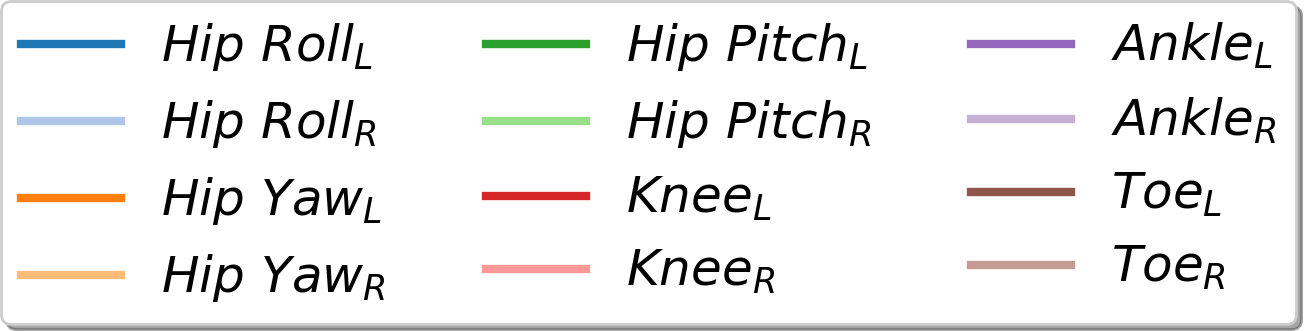}
	\caption{Demonstration of the impact invariant projection on joint velocity data from a successful jumping experiment on the physical Cassie robot. Joint velocities (top) during the landing event change rapidly which is difficult to perform feedback control on. By projecting the same joint velocities to the impact invariant subspace (bottom), the values are more consistent and more amenable for feedback control. Note, the change in joint velocities primarily occurs within a time span of only 10 ms. The L and R subscripts indicate the left and right leg respectively.}
	\label{fig:hardware_joint_vels}
\end{figure}

To illustrate the benefit on a \textit{physical robot}, we apply the impact invariant projection to the joint velocities for Cassie executing a jumping motion right when it lands as shown in Fig. \ref{fig:hardware_joint_vels}.
Observe that the projected joint velocities are significantly smoother than the original joint velocities.
\subsection{Application to Joint Space Tracking}
	
	Many joint space controllers have been formulated for the control of bipedal robots. 
	These include controller such as Hybrid LQR \cite{mason2016balancing}, joint PD control \cite{gong2019feedback} \cite{green2020learning}, as well as inverse dynamics controllers that primarily track joint space outputs \Will{ \sout{\cite{gong2020angular},}} \cite{reher2020inverse}.
	Utilizing the impact invariant projection on these controllers is straightforward.
	In a small time window around the anticipated impact, simply replace the original control law 
	\begin{align*}
		u = u_{ff} + K_p \tilde{q} + K_d \dot{\tilde{q}},
	\end{align*}
	with the new projected joint velocity error, which results in
	\begin{align}
		u^* = u_{ff} + K_p \tilde{q} + K_d Q(q) \dot{\tilde{q}}.
	\end{align}

\subsection{Application to Task Space Tracking}
	\subsubsection{Operational Space Controller}
	
	When the tracking objectives are instead more general functions of the robot's state, this style of controller is commonly referred to as operational space control (OSC).
	An OSC is an inverse dynamics controller that tracks a set of task or output space accelerations by solving for dynamically consistent inputs, ground reaction forces, and generalized accelerations \cite{wensing2013generation}
	\Will{\cite{sentis2005control}}.
	For an output position $y(q) = \phi(q)$ and corresponding output velocity $\dot y = J_y(q) \dot q$, where $J_y(q) = \PartialDiff{\phi}{q}$,
	the commanded output accelerations $\ddot {y}_{cmd}$ are calculated from the feedforward reference accelerations $\ddot y_{des}$ with PD feedback:
	\begin{align}
	\ddot {y}_{cmd} &= \ddot{y}_{des} + K_p (y_{des} - y) - K_d (\dot{y}_{des} - \dot y)
	\label{eq:output_cmd}
	\end{align}
	The objective of the OSC is then to produce dynamically feasible output accelerations $\ddot y$ given by: 
	\begin{align*}
	\ddot y &= \dot J_y \dot q + J_y \ddot{q},
	\end{align*}
	such that the instantaneous output accelerations of the robot are as close to the commanded output accelerations as possible.
	This controller objective can be nicely formulated as a quadratic program:
	\begin{align}
	\min_{u, \lambda, \ddot q}   &\quad&  \sum_{i}^{N} ({\ddot y_i - \ddot y_{i_{cmd}}})^T W_i (\ddot y_i - \ddot y_{i_{cmd}}) &&          &  \label{eq:osc_qp}\\
	\text{subject to: }
	&\quad&  \text{Dynamic Constraints}  &&  & \label{eq:dyn_constraint} \\ 
	&\quad&  \text{Holonomic Constraints}  &&  & \label{eq:constraint_satisfaction} \\ 
	&\quad&  \text{Friction Cone Constraints}. &&  & \label{eq:push_constraint}
	\end{align}
	$i$ denotes the particular output being tracked (e.g., center of mass or foot position) and $W_i$ are corresponding weights on the tracking objectives.
\subsubsection{Projecting Outputs to the Impact Invariant Subspace}

	The desired and actual positional outputs $y_{des}$ and $y$ are trivially continuous over impacts.
	Thus, the impact invariant projection is applied only to the output velocities $\dot y$ and $\dot y_{des}$.
	However, due to the lack of a one-to-one mapping between output velocities $\dot y$ and generalized velocities $\dot q$, the projection in \eqref{eq:ii_projection} cannot be naively applied.
	Still, it is possible to project the output velocity to a subspace so that it is invariant to any unknown contact impulse.
	In practice, this can be accomplished for a single output with the following optimization problem:
	\begin{align}
		\min_{\lambda} \quad \TwoNorm{\dot y_{des} - J_y(\Will{\dot q} + M^{-1}J_{\lambda}^T \lambda)}.
		\label{eq:ii_optimization_problem}
	\end{align}
	This applies a correction to the generalized velocities $\dot q$ that minimizes the tracking error in the output velocities, under the condition that the correction lies within the set of feasible velocities that could result from a contact impulse $\lambda$.
	In the absence of constraints on $\lambda$, this can be formulated as a least squares problem and the optimal $\lambda$ can be solved for implicitly with the Moore-Penrose pseudo-inverse denoted by $(\cdot)^{\dagger}$.
	The projected output velocity error, $\dot{y}_{proj}$, can then be found as:
	\begin{align}
		\dot{y}_{proj} &= \dot y_{des} - J_y \dot q - J_y \dot q_{\lambda},
		\label{eq:ii_osc_projection}
	\end{align}
	where the correction $\dot q_{\lambda}$ is given by:
	\begin{align}
		\dot q_{\lambda} &= M^{-1} J_{\lambda}^T(J_y M^{-1} J_{\lambda}^T)^{\dagger}(\dot y_{des} - J_y \dot q).
		\label{eq:vel_proj}
	\end{align}
	This projected error $\dot y_{proj}$ is then used in place of the original output velocity error $\dot y_{des} - \dot y$ in \eqref{eq:output_cmd}.
	This can have two interpretations.
	One interpretation, which is more literal, is that we apply a correction in the space of $\lambda$ that minimizes the velocity tracking error in the output space.
	The other interpretation, is that the correction projects the velocity tracking error to the impact invariant subspace by eliminating the sensitivity of the error on $\lambda$.
	In either interpretation, it is easy to see that $\dot y_{proj}$ is invariant to any unknown contact impulse.
	Note, the solution to the least squares problem, $\lambda$, is not intended to be most physically plausible contact impulse, but instead the impulse that minimizes the tracking error.
	In both interpretations, the projection assumes an \textit{optimistic} correction in the $\lambda$ space.
	\Will{Related work has also explored incorporating potential impacts into robust control formulations \cite{wang2019impact}. A primary distinction is that \cite{wang2019impact} seeks to be robust to impulsive impacts from a known model but at unexpected times, where our approach makes no assumptions on the magnitude or duration of impact forces.}
	
	\remark{\textit{$\dot q_{\lambda}$, defined in \eqref{eq:vel_proj}, can easily be constructed when there are multiple tracking objectives by simply stacking the output space Jacobians and velocity errors.
		The derivative gains and weighting matrices defined in \eqref{eq:output_cmd} \eqref{eq:osc_qp} can similarly be included, but in practice we did not find a noticeable effect from including them.}
	}

	For output spaces like the position of impacting foot, $\dot y_{proj}$ is exactly zero.
	This can be seen because $J_y$ is identical to $J_{\lambda}$, and thus \eqref{eq:ii_osc_projection} equals 0 for all states.
	Intuitively, this makes sense because the foot undergoes large changes in its velocity when it rapidly comes to rest when it makes contact with the ground, and thus we should not attempt to control the foot velocity during impact.
	Therefore, for some outputs, the projection behaves similar to applying no derivative feedback.
	
\subsubsection{Constraints on the Projection}
	
	Due to the lack of constraints on $\lambda$, the projection impulse is not guaranteed to be physically possible.
	$\lambda$ could be constrained to lie within the friction cone $FC$.
	However, upon further examination, inclusion of these constraints may be undesirable.
	This is because sensitivity to the impact event can result from the \textit{absence} of expected impacts as well - consider the case when the robot makes contact after the nominal impact time.
	Constraining $\lambda \in FC \cup -FC$ is not practical as this set is non-convex.
	Although it is possible to formulate this as a binary mixed integer program, solving this problem was considered to require too many assumptions to justify the additional complexity.

\section{Evaluation}
\label{sec:evaluation}

To showcase the advantages of using the impact invariant subspace, we apply the aforementioned projection on two examples in simulation: a walking controller for the planar five-link biped Rabbit \cite{chevallereau2003rabbit} and a jumping controller for the 3D bipedal robot Cassie. 
We then adapt the jumping controller to the physical robot Cassie and show that the we can achieve similar effects on hardware.

%

\subsection{Controller Details}

\subsubsection{Finite State Machine}
We use a time-based finite state machine (FSM) to transition between the tracking objectives and the active contact mode of the OSC.
\sout{Though event-based finite state machines with contact detection as the trigger have been developed for legged robots \cite{bledt2018contact}, detection still has delays of 4-5 ms at best and does not address the other uncertainties present in the impact event.}
In a small time window of duration $T$ before and after the nominal impact time, we blend in the correction $\dot q_{\lambda}$ using a continuous scalar function $\alpha(t)$ to avoid introducing additional discontinuities.
The scalar function is given by:
\begin{align}
	\alpha(t) = 1 - \exp(\frac{-(t - t_{switch} + T)}{\tau}),
\end{align}
where $t_{switch}$ is the nominal impact time given by the reference trajectory, and $\tau$ is the time constant.
We then use $\alpha(t)$ to modify \eqref{eq:ii_osc_projection} to be:
\begin{align*}
	\dot{y}_{proj} &= \dot y_{des} - J_y \dot q - J_y (\alpha(t) \dot q_{\lambda}).
\end{align*}
Note our choice for the blending function $\alpha(t)$ is arbitrary; any monotonic continuous function with a range $\in [0, 1]$ can accomplish a similar purpose.



\subsubsection{Reference Trajectories}
\label{subsec:reference_trajectories}

The target walking and jumping trajectories were solved for offline by solving a constrained trajectory optimization problem on the respective full order models.
The problems were transcribed using DIRCON \cite{posa2016optimization} and solved using SNOPT \cite{gill2005snopt}.
The jumping trajectory was constrained to have the robot pelvis reach an apex height of 15cm above its initial starting height and to have both feet have 15cm of clearance from the ground at the apex.
The same jumping trajectory was used in both simulation and on the physical robot. 

\subsection{Joint Space Controller for Rabbit Walking}

\subsubsection{Experimental Setup}
To demonstrate the directional nature of the impact invariant projection, we apply the projection to a \Will{joint space inverse dynamics} \sout{OSC} walking controller for the planar biped Rabbit in simulation with the hip and knee joint angles in both legs as the outputs.
We generate a periodic walking trajectory using trajectory optimization and perturb the swing foot vertical velocity by 0.1 m/s at the start of the trajectory so that the robot makes contact away from the nominal impact time.
To evaluate robustness, we compare the post-impact velocity error in the joints of both the swing and stance leg for three variations of the joint space controller:
\begin{itemize}
	\item No adjustment: this is a standard joint space controller that makes no special considerations with regards to the impact event other than switching contact modes at the nominal impact time.
	\item No derivative feedback ($K_d = 0$) applied in a window 25ms before and after the nominal impact time.
	\item Impact invariant projection applied in a window 25ms before and after the nominal impact time.
\end{itemize}

We add the additional comparison to a controller with no derivative feedback to demonstrate the structure of the impact invariant subspace.
While both applying no derivative feedback and using the projection seek to reduce the sensitivity to large velocity errors at the impact event, the projection solves this problem in a more principled fashion that leads to better tracking performance.
Note, no direct comparison can be made to \cite{gong2020angular}. 
Although we could regulate $L$ alone, in \cite{gong2020angular} the walking controller regulates $L$ through footstep planning, which occurs at a much slower frequency than regulation using motor torque inputs.

\subsubsection{Results}

\begin{figure}
	\centering
	\includegraphics[width=0.48\textwidth]{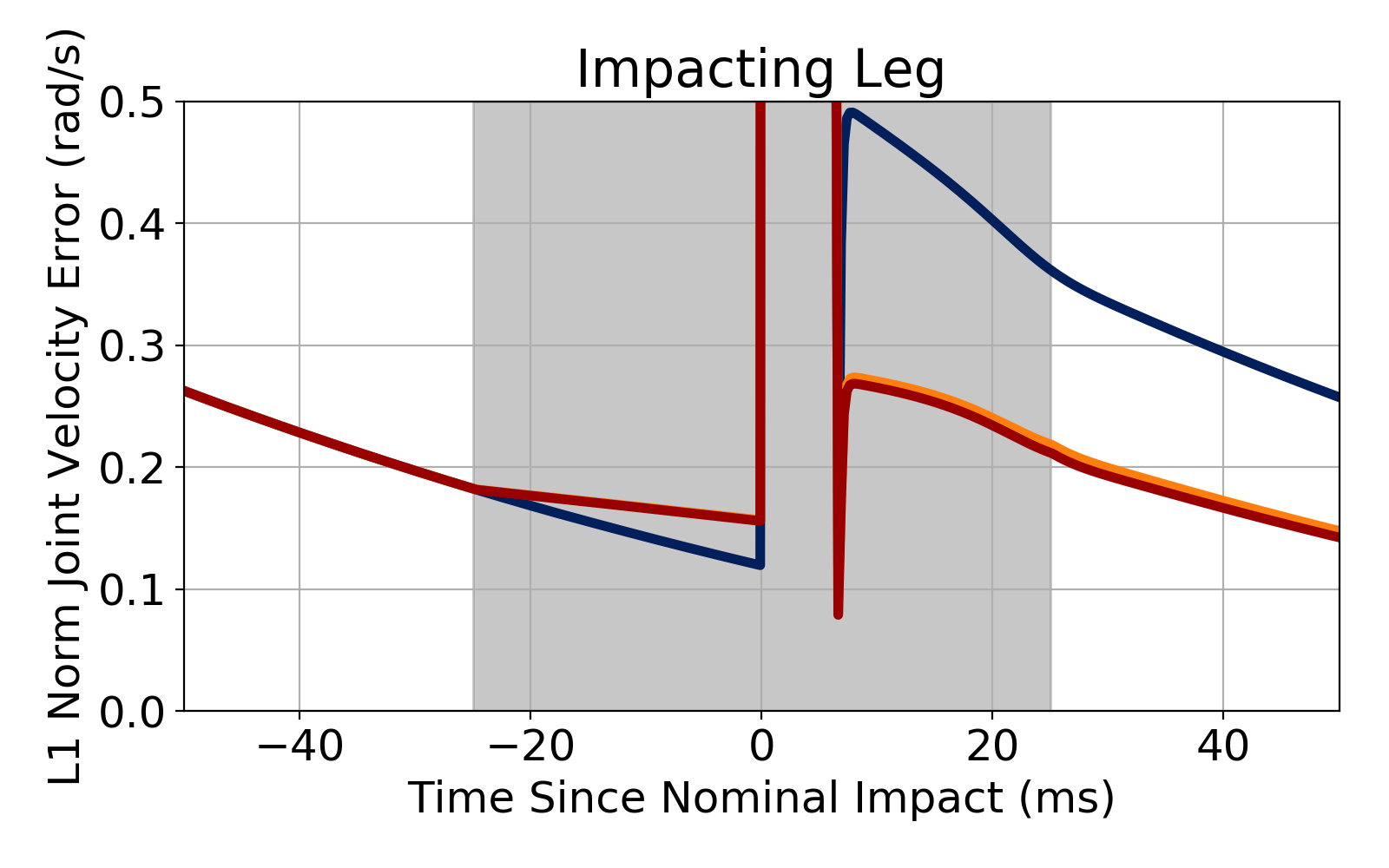}
	\includegraphics[width=0.48\textwidth]{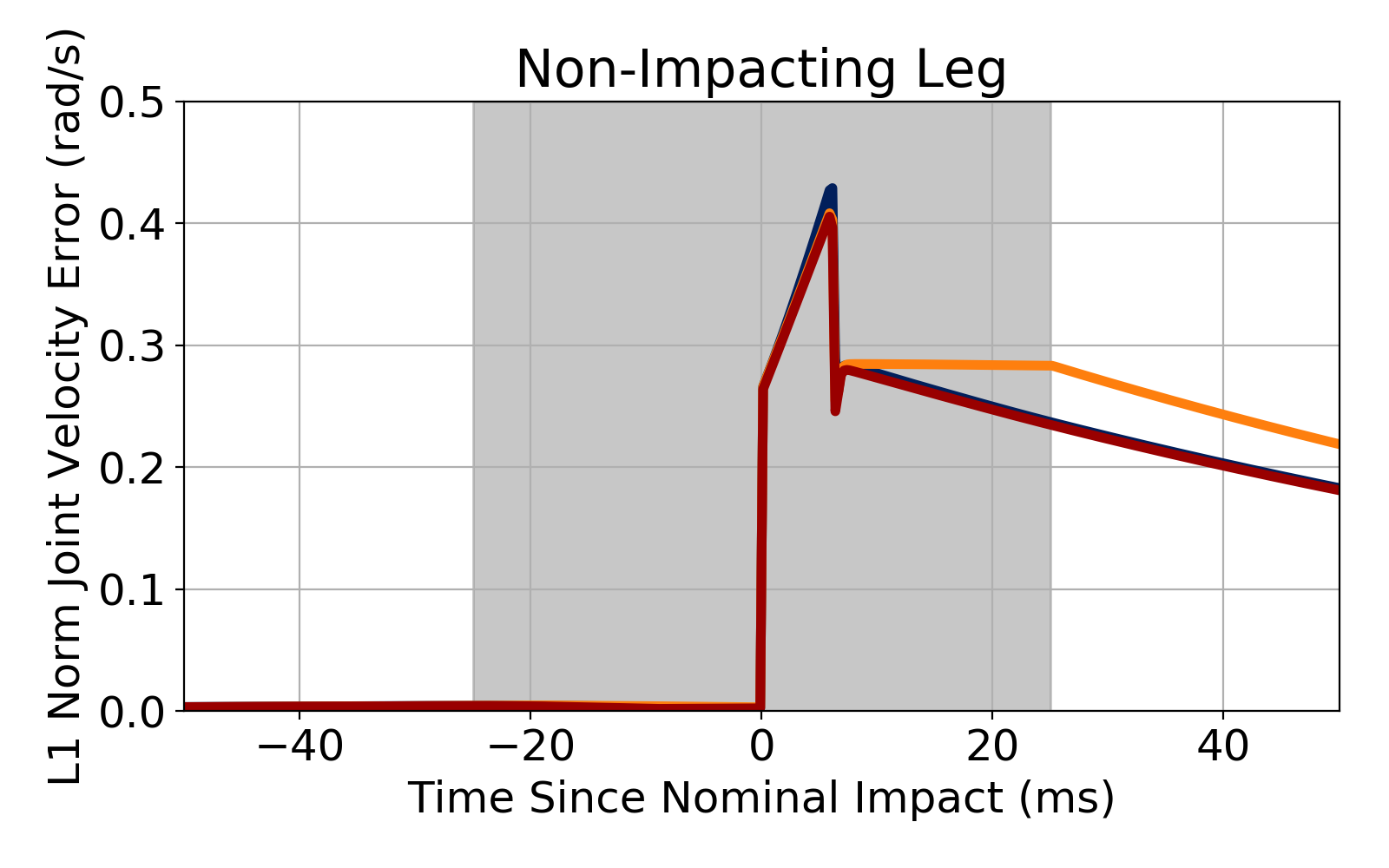}
	\includegraphics[width=0.48\textwidth]{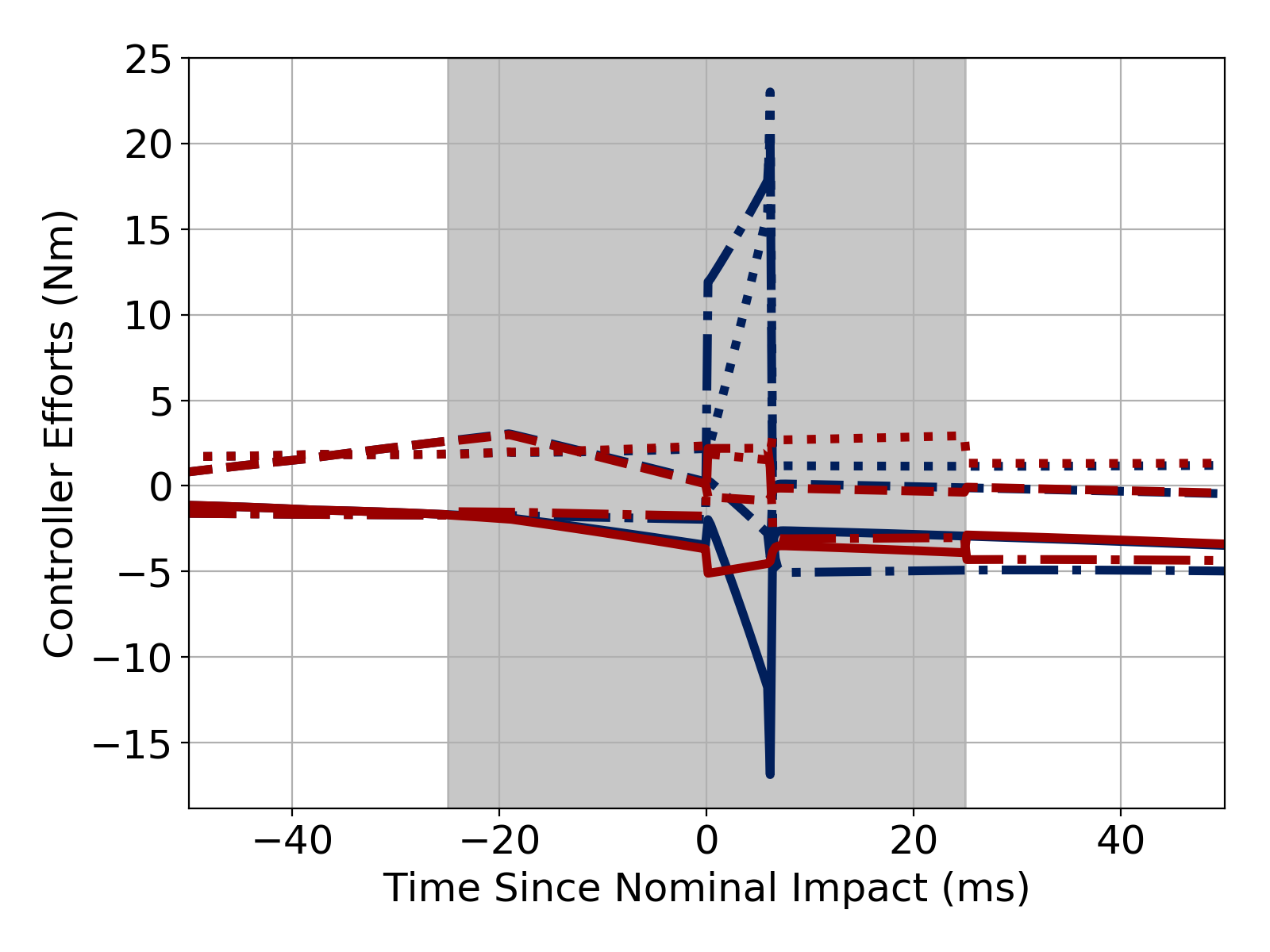}
	\includegraphics[width=0.46\textwidth]{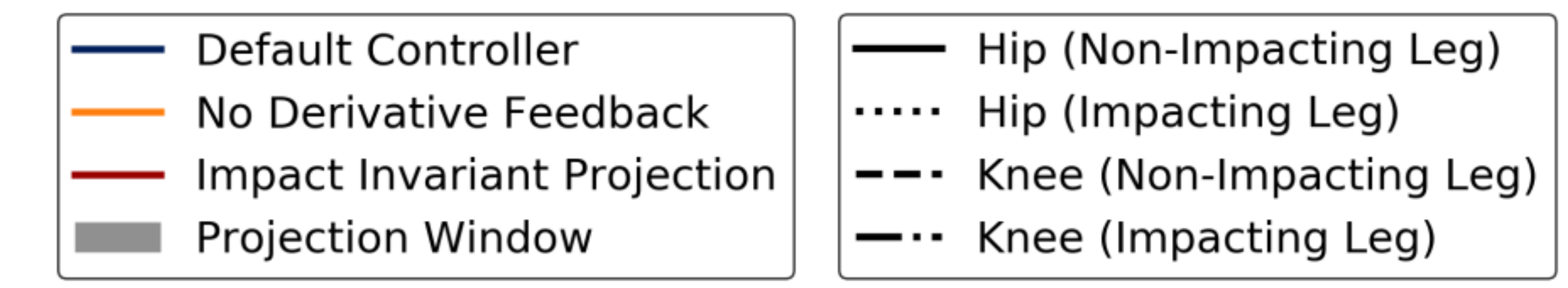}
	\caption{The joint velocity tracking errors are shown for the swing (impacting) leg (top) and the stance (non-impacting) leg (middle) for all three control strategies. The controller that utilizes the impact invariant projection is shown to be robust to the mismatch in impact timing as evidenced by lower tracking error compared to the default controller. The impact invariant controller is also able to maintain full control authority over the joints in the non-impacting leg compared to the controller that applies no derivative feedback in the same window. Note, the better performance of the impact invariant controller is because it is successfully ignores velocity errors due to the impact event, which is evident from the applied controller efforts (bottom). We omit the controller efforts from the no derivative feedback controller to reduce clutter.}
	\label{fig:directional_robustness}
\end{figure}

Shown in Fig. \ref{fig:directional_robustness}, the controller using the impact invariant projection is able to achieve the best tracking performance out of the three controllers for \textit{both} legs.
It has better tracking performance than the default controller for the joint velocities of the impacting leg by not overreacting to the impact event as shown by the controller efforts.
At the same time, it has better tracking performance than the controller with no derivative feedback for the joint velocities of the non-impacting leg by appropriately regulating the velocities in those joints.

\subsection{Operational Space Jumping Controller}
\subsubsection{Experimental Setup}

Next, we evaluate the performance of the impact invariant projection on a jumping controller for Cassie.
We chose to look at jumping due to the richness of the impact event: the robot cannot accurately estimate its state when it is in the air and must make impact with the ground with non-zero velocity.
We set the target outputs of the OSC to be the position and orientation of the pelvis during the initial crouching phase. 
We then switch to track the feet positions in flight due to the uncontrollability of the center of mass. 
At the nominal landing time, we switch back to tracking the position and orientation of the pelvis. 
Similar outputs are used in another jumping controller for Cassie \cite{xiong2018bipedal}.

To measure the robustness to uncertainty in the impact event, we perturb the system in the Drake \cite{drake} simulator by introducing a platform of differing heights $[0cm, 5cm]$ at the final landing location and by adjusting the penetration allowance parameter $\in \{10^{-5}, 10^{-4}, 10^{-3}, 5*10^{-3}\}$ of the contact model used by the simulator.
Note, the penetration allowance parameter is roughly equal to the maximum interpenetration between the foot and the ground in meters; though for high-impact motions such as jumping, the maximum penetration can be 2-3 times that.
The terrain height effectively changes the impact time while the penetration allowance adjusts the stiffness of the ground and therefore the ground reaction forces during impact.
For details on the contact model used in the time-stepping Drake simulator, consult \cite{castro2020transition}.
Additionally, we adjust the duration of the projection window to evaluate the sensitivity of the controller performance to that parameter.
We evaluate the controller performance on two metrics, the control effort and the acceleration error at the end of the impact event.

\begin{itemize}[leftmargin=*]
	\item Control Effort:
		We quantify the control effort used to stabilize the robot upon landing using the objective:
		\begin{align*}
			J_{mot} = \int_{t_0}^{t_f} u^2 dt,
		\end{align*}
		where $t_0$ and $t_f$ are the start and end of the projection window respectively.
		We use the projection window with the maximum duration of $50ms$ to ensure a fair comparison; this is because all the controllers are identical outside of this time window. 
		We choose this as a measure for how hard the actuators are working, with the idea that minimal control effort is desirable, especially when applying feedback control when the state is so uncertain.
	\item Acceleration Error:
		We used the weighted acceleration error in the output space to measure the tracking error after impact:
		\begin{align*}
			\ddot y_{err} &= K_p (y_{des} - y) - K_d (\dot{y}_{des} - \dot y),\\
			J_{acc} &= \sum_{i = 0}^{N} \ddot y_{i, err}^T W_i \ddot y_{i, err},
		\end{align*}
		where $N=2$ because during the landing phase we track only the pelvis position and orientation.
		We choose this as a way to combine position and velocity error into a single metric.
		We evaluate the acceleration error at end of the longest projection window because after this point, all the controllers are identical.
		To avoid bias from sampling at a particular time step, we sample multiple points near the desired sample time.
		To put the acceleration error in context, we normalize the acceleration error so that an acceleration error of $1.0$ corresponds approximately to a tracking error of $7cm$ for the vertical position of the pelvis.
\end{itemize}

\subsubsection{Results}
\begin{figure}
	\centering
	\includegraphics[width=0.48\textwidth]{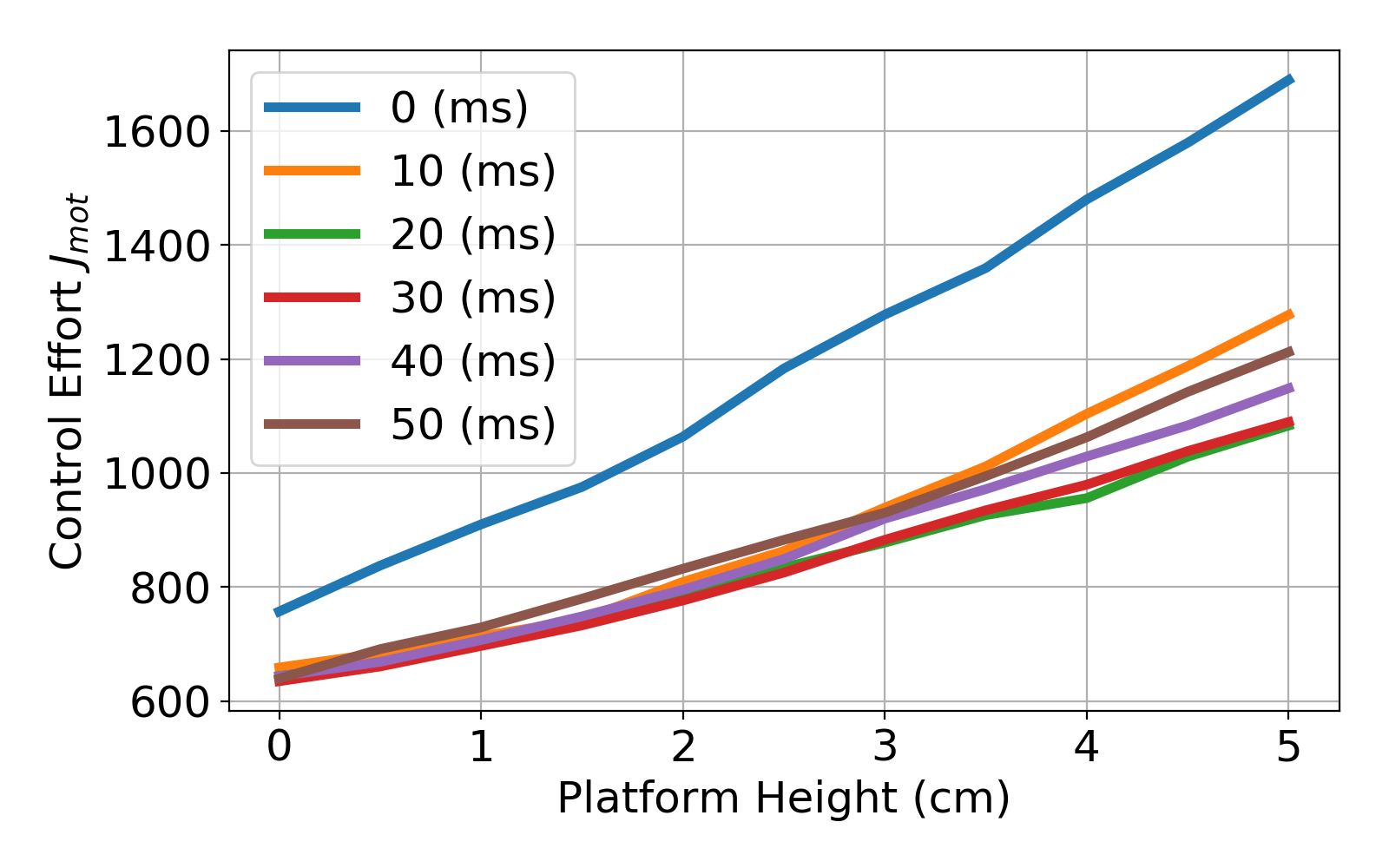}
	\includegraphics[width=0.48\textwidth]{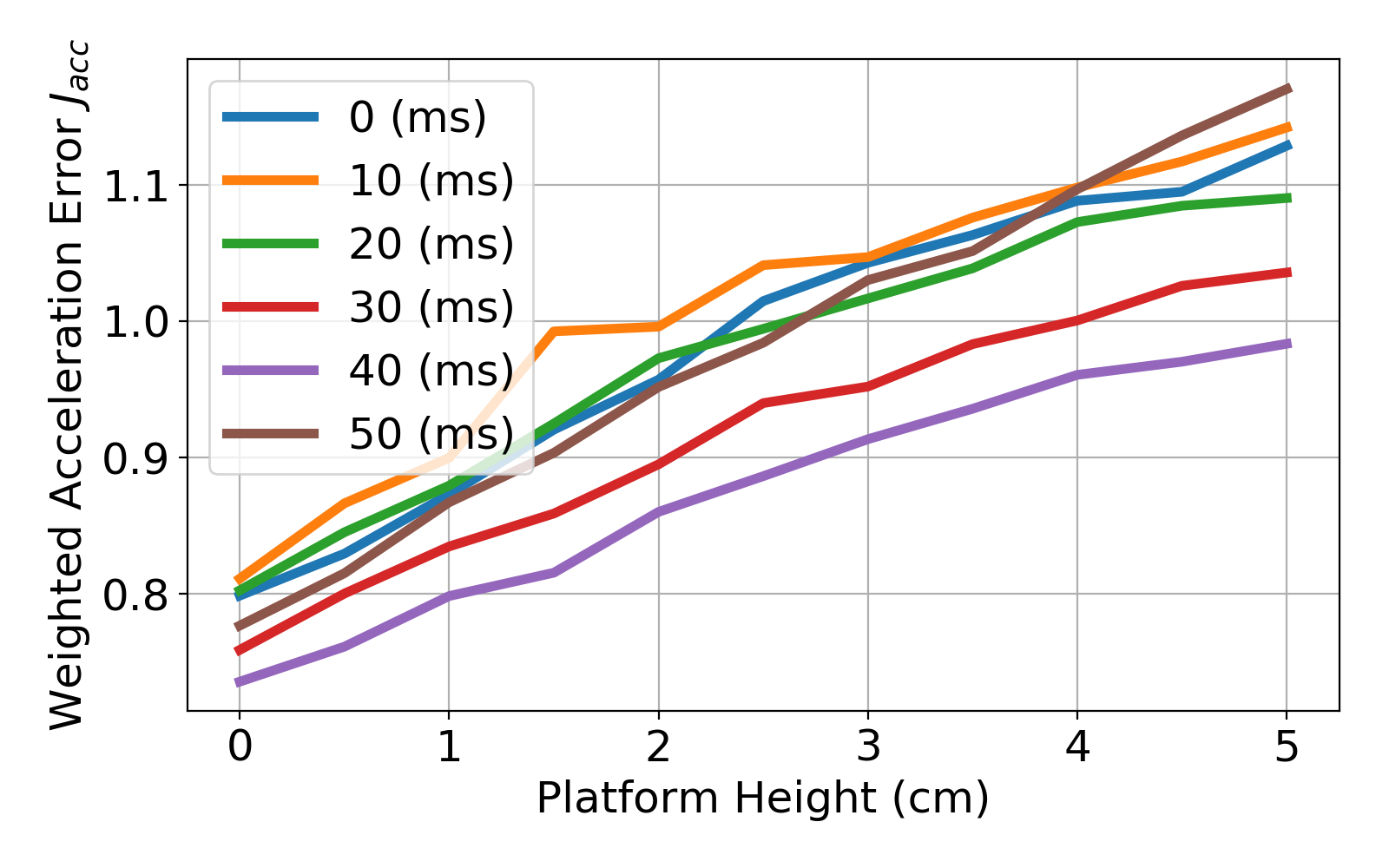}
	\includegraphics[width=0.48\textwidth]{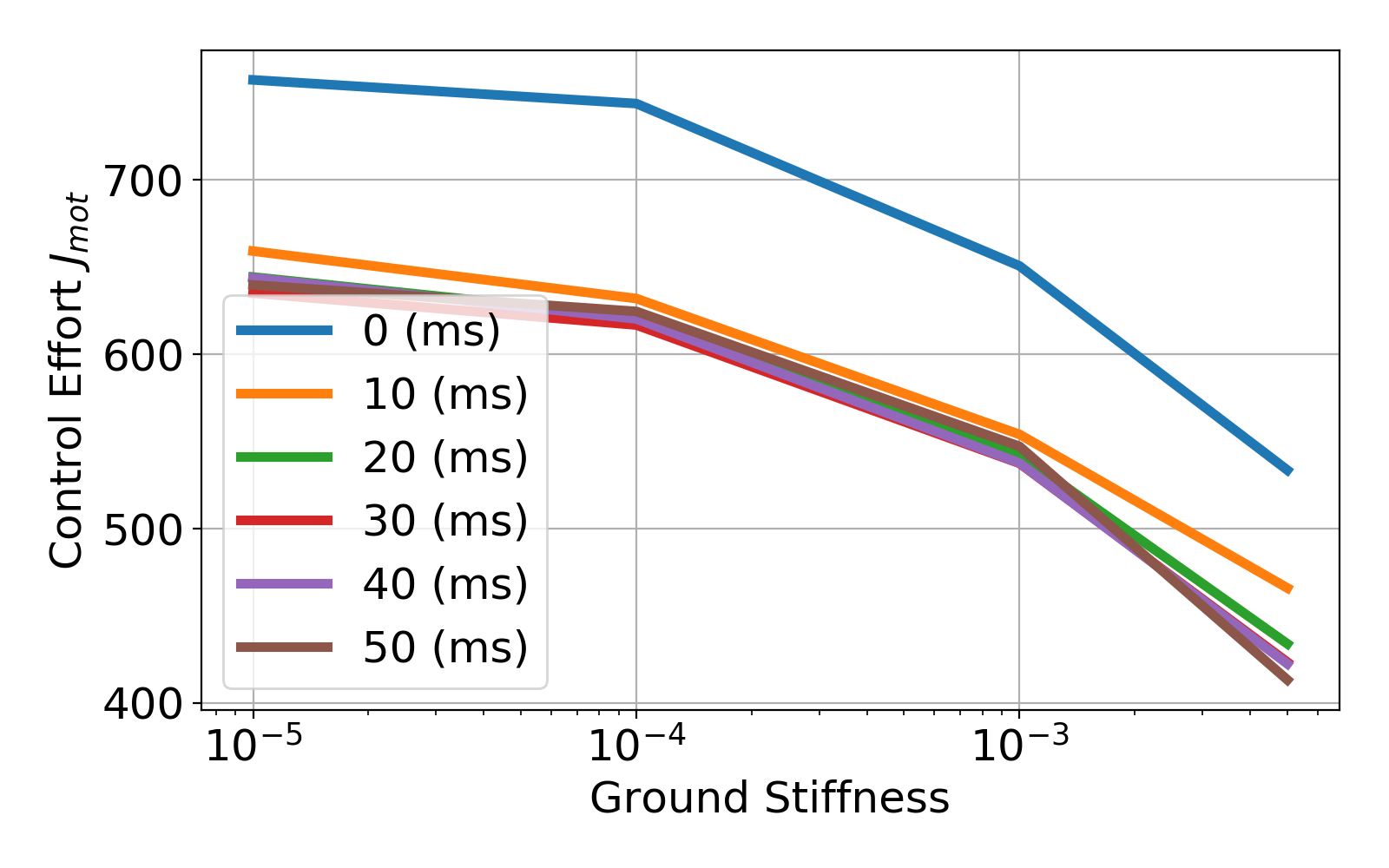}
	\includegraphics[width=0.48\textwidth]{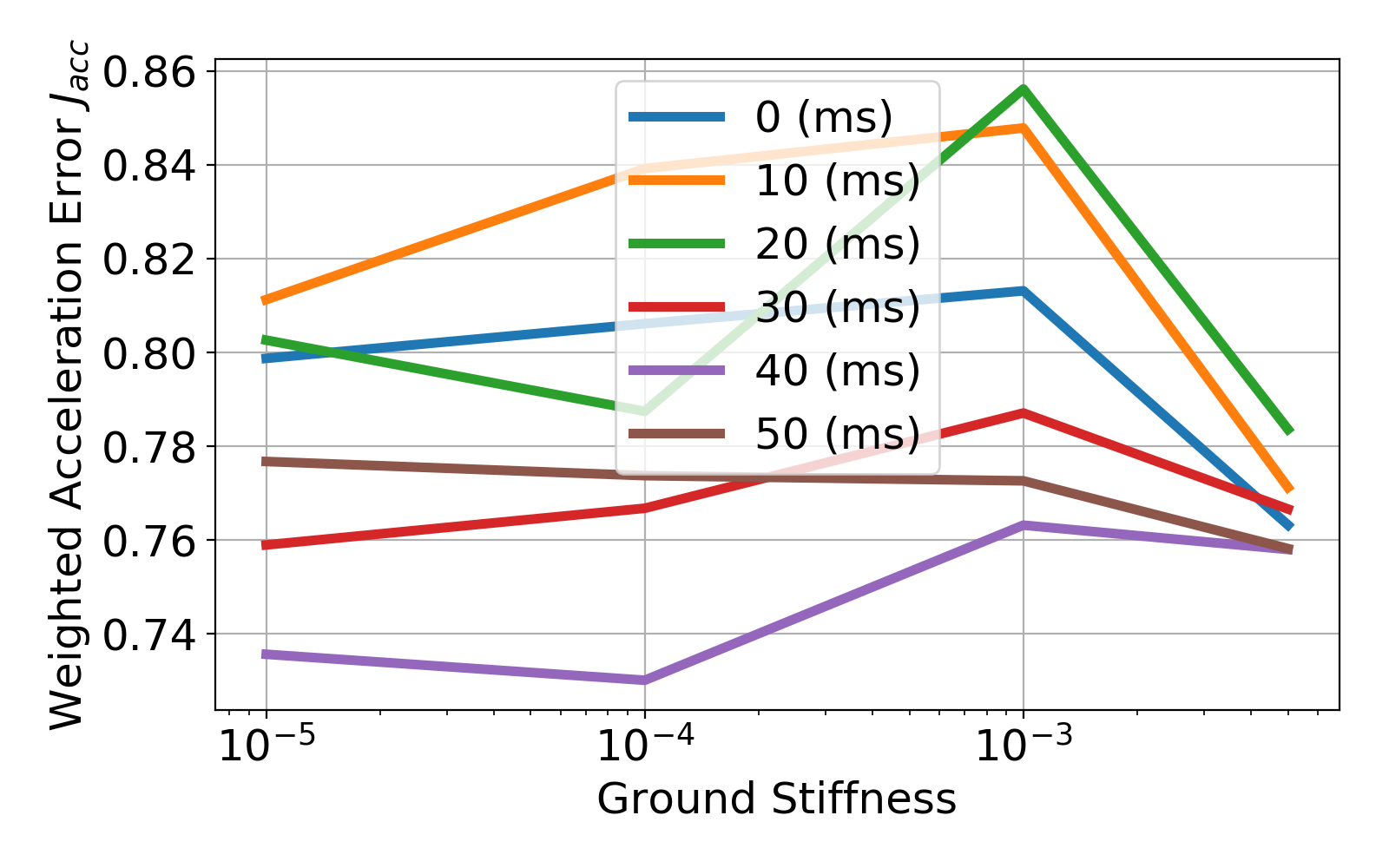}
	\caption{Results from the parameter sweep over unexpected landing heights and ground stiffnesses \Will{for varying durations of the impact invariant projection}. The controllers with the impact invariant projection have a significantly \Will{larger} control effort cost compared to the default controller across all terrain heights and stiffnesses. The projection also reduces the weighted acceleration error, though the benefit is sensitive to the duration of the projection.}
	\label{fig:param_study_results}
\end{figure}

The results from the sweep over changes in the landing platform height and ground stiffness are shown in Fig. \ref{fig:param_study_results}.
\sout{In the control effort, we see a distinct reduction across all perturbations if we apply the projection for any of the tested durations.
We observe less sensitivity to the perturbation to the platform height as well, thus showcasing the improved robustness to uncertainty in the impact event.}
\Will{We see a distinct reduction in the absolute control effort across all perturbations when applying the projection for all of the tested durations.
We observe less sensitivity to the perturbation to the platform height as well, showcasing the improved robustness to uncertainty in the impact event.}
We see a noticeable improvement in the acceleration error for perturbations in platform height and ground stiffness as well, although the performance of the controllers on this metric is significantly more sensitive to the duration \Will{of the projection.}
\sout{for when the projection is applied}
A somewhat surprising result from the sweep across ground stiffnesses is the lower cost in control effort for the softer ground.
This reduction in control effort is due to more gradual changes in the velocities at impact, which results in lower peak actuator efforts - something that is penalized in our cost function.
The acceleration error is mostly constant across ground stiffnesses and shows a similar benefit for using \Will{the} projection.

\subsection{Experiments with the Physical Robot}
Finally, we apply the impact invariant projection on the physical robot to evaluate the robustness to actual uncertainties.
Implementing the jumping controller described above onto Cassie requires some minor additions discussed here.

	\begin{itemize}[leftmargin=*]
		\item State Estimator: 
		We use the contact-aided invariant EKF developed in \cite{hartley2020contact} to estimate the floating-base pelvis state.
		\item Contact Estimation: 
		Although we do not directly use contact detection in our controller, the state estimator utilizes contact data in the measurement update.
		Cassie does not have dedicated contact sensors, but contact detection can be achieved through proprioception.
		We use a generalized-momentum observer, similar to the method used in \cite{bledt2018contact}, to estimate the contact force at each foot.
		We then set a threshold of 60 Nm on the contact normal force to define contact.
		We observe that this has a faster response and better accuracy over detecting contact using spring deflections.
		\item Output Trajectory Adjustments:
		Due to the lack of accurate estimation of the robot's state in the global frame, particularly in the flight phase, we express the reference output trajectories in robot local coordinates.
		The pelvis trajectory for both the crouching and landing phases is defined with respect to the center of the support polygon, and the swing feet trajectories are defined with respect to the corresponding hips.		
	\end{itemize}

	\subsubsection{Experimental Results}
	
\begin{figure}
	\centering
	\includegraphics[width=0.48\textwidth]{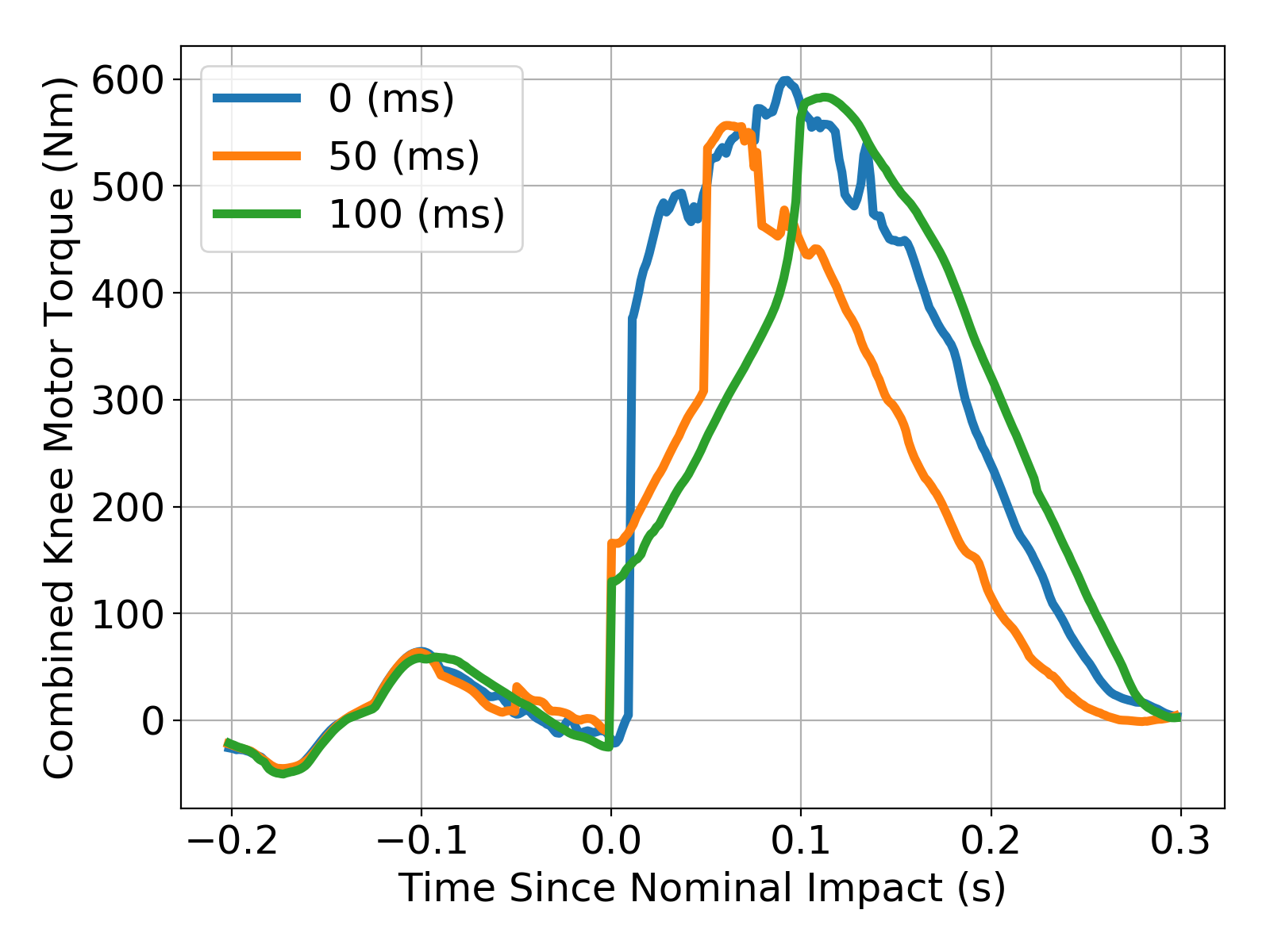}
	\caption{Motor efforts on Cassie executing the jumping motion. The combined knee motor torques commanded by the jumping controller are shown for three different durations of the impact invariant projection. Notice that the experiments using the controller with the projection are able to avoid the initial spike in controller efforts at the impact event. This enables the controller to respond instead to the velocities after the \sout{the} impact has resolved.}
	\label{fig:hardware_knee_efforts}
\end{figure}

	Experiments using the controller with and without the impact invariant projection were both consistently able to successfully complete the jump.
	Snapshots of the jumping motion are shown in Fig. \ref{fig:cassie_hardware_jump}.
	As seen in Fig. \ref{fig:hardware_joint_vels}, the joint velocities change rapidly during the impact event.
	By projecting the velocity error of the outputs (position and orientation of the pelvis) to the impact invariant subspace, we avoid overreacting to these rapid velocity changes in a principled manner.
	The effects of this can be seen in Fig. \ref{fig:hardware_knee_efforts}, where the \Will{ \sout{peak} change in} knee motor efforts, particularly at the impact event, are significantly reduced when using the impact invariant projection.
	We choose to show the knee motor efforts because they exhibit the largest change at the impact event due to their role in absorbing the weight of the pelvis at impact.
	This smoothing out of the commanded motor efforts is similar to what we see in simulation.
	The jumping motions for the controller both with and without the impact invariant projection are also shown in the supplementary video.

\section{Discussion}
\label{sec:discussion}

In this paper, we introduce a general framework that enables controllers to be robust to uncertainties in the impact event without sacrificing control authority over unaffected dimensions.
We achieve this by projecting the control objectives to an impact invariant subspace.
Using the impact invariant projection on the physical robot, we are able to prevent the controller from reacting to mismatches in impact timing and reduce undesirable spikes in the controller efforts.
We believe that this will aid in developing controllers for legged robots that are capable of more dynamic motions such as running.

As the parameter studies and experiments on the physical robot show, the controller performance is slightly sensitive to the duration of the impact invariant projection and will therefore have to be tuned.
However, tuning the duration in practice is straightforward; applying too short of a window may miss the impact event and applying the window for too long hinders the overall tracking performance.

While this work focuses on discontinuous in the reference trajectories and the robot's state, we acknowledge that discontinuities in the controllers of legged robots can arise from other sources as well.
For example, the jumping controller switches from tracking the feet to tracking the pelvis after it lands. 
At the same time, the active contact mode changes from no contact to double stance.
It is not clear whether these discontinuities are avoidable or even problematic; still, a careful exploration into the entirety of controller decisions around an impact event is warranted.

Finally, although this framework is robust to any possible contact impulse, this may ultimately be too heavy handed.
Instead, by restricting and identifying the uncertainties present during an impact event, we can more carefully construct our controllers to respond to those uncertainties rather than ignoring them.
Future work therefore includes identifying the uncertainty parameters that are relevant to our controllers, including the controller decisions identified above.

\section*{Acknowledgements}

We thank Yu-Ming Chen and Brian Acosta for help with hardware testing.

\bibliographystyle{IEEEtran}
\bibliography{references.bib}

\end{document}